# Expert Systems with Logic#
# A Novel Modeling Framework for Logic Programming in an Object-Oriented Context of C#

Fabian Lorenz[1]    Maik Günther[2]

**Abstract:** A good choice of the programming paradigm can reduce the required development effort. Nevertheless, today almost all industrial systems are developed using imperative or object-oriented approaches. In order to use the advantages of other programming paradigms meanwhile these approaches are implemented into common imperative and object-oriented languages. While this is the standard for the functional paradigm in many languages today, logical programming is hardly supported, although there are practical applications here as well, such as expert systems. This paper presents a novel approach how logical programming can be declared directly in an object-oriented language. Also, some programming examples are shown using an implementation of the approach in C# called Logic#. Furthermore, these examples will be used to evaluate that the approach helps to develop programs better than with C# in its native version, but also points out that further improvements are still possible as the native logic language Prolog shows even better results.

**Keywords:** expert systems, logic programming, object orientation, C#, Prolog, Logic#, modeling framework

## 1 Introduction

With the development of the programming languages in the course of the 20th century different approaches were developed, how a program can be described [1]. A central difference lies above all whether one describes how the program sequence is to proceed or instead the problem definition must be declared as exactly as possible, so that the computer can compile the necessary computations itself [2]. In the today's industrial software development, thereby above all the first approach and thus the imperative as well as the object-oriented paradigm based on it is used. However, it could be evaluated experimentally that the selection of an approach ideally suited to the task enables development with less scope and source code complexity [3]. For example, it is shown that a synergy arises in the development of expert systems with logical languages based on the syntactic similarity of logic programs and the way the human brain conceives of such logical programs [4]. Such examples show that the use of the widely used imperative languages is not always optimal and that other languages would sometimes provide better results. That these languages are nevertheless mostly chosen can best be explained by the fact that the imperative language family and its offshoots are taught in school and university courses as primary programming languages [5]. Nevertheless, in order to gain advantages from the various foundational principles of programming, approaches from other paradigms such as functional programming are ported in widely used imperative languages [2]. However, the situation is different for the logical programming mentioned earlier, which is typically represented by the Prolog language. There are Prolog interpreters for many languages which can be attached as third-party-tool, however no usual language makes the native development of logic possible [6].

[1] IU International University of Applied Sciences, Juri-Gagarin-Ring 152, 99084 Erfurt, Germany, fabian.lorenz@iubh-fernstudium.de
[2] IU International University of Applied Sciences, Juri-Gagarin-Ring 152, 99084 Erfurt, Germany, maik.guenther@iu.org

Alongside this paper, a novel approach will be introduced, how logic programming can be implemented in an object-oriented language context. To further illustrate this concept, we will look at Logic#, an implementation of this approach in C# [7]. Both, the definition of the syntax, with which logic programs can be declared in object orientation, and the algorithms, were inspired by Prolog. Be aware that the introduced approach will not be able to introduce all the features of logical programming. However, it should show how the central approach of logical programming can be used with Object-orientation. In the further course of the paper, it will be evaluated whether there are any improvements at all for programmers by using a framework that implements this approach. In particular, the presented approach will be compared with both native C# and native Prolog.

This paper is structured as follows: In chapter two, the architecture of a logic framework in an object-oriented environment is presented. In the following third chapter Logic# is considered based on example programs. Chapter four then addresses the question of whether the approach is useful at all and compares it with native C# and native Prolog for this purpose. This paper end with a discussion of the results and a conclusion.

## 2 Software Architecture of Logic#

The base for this approach is an architectural design, which can then be used for implementation in various object-oriented languages. There are two aspects that must be considered for the architectural design. First of all, a class model is required with which a user can declare logic programs and also query them. On the other hand, algorithms are needed to solve these queries.

We will start with the study of the requirements for the definition of logic programs. As already mentioned, the architecture is inspired by Prolog, so as a first step it is considered which syntax components a Prolog program contains and thus must be represented in the class model. In Tab. 1 the necessary components are noted. At this point it is to be pointed out again that not all components of Prolog were taken over, but was limited to the central declaration of the knowledge base by facts and rules. Thus, arithmetic operations and list processing are not integrated in Logic# as they are already available in the default scope of functions in C#.

Facts as well as rules are necessary as inputs, which contain in each case a predicate with an indefinite number of attributes, whereby the attributes are represented by objects. In addition, the rules require a set of conditions, which can be logically linked with "AND" as well as "OR". It is noticeable that the first part is the same for rules and facts, but this can be attributed to the fact that both components are represented by horn clauses, which may or may not have a body depending on their expression [8]. Comparable to Horn clauses, both requirements can thus be unified by representing facts as rules whose condition part is empty. Attributes need the possibility to represent a variable instead of a constant object. For this, the type of the object should be decisive, which is why a class for encapsulating variables is necessary for this. The last requirement for the inputs is the creation of queries, which should consist of a predicate. For this concept, we will limit ourselves to simple, unlinked queries. The definition of variables can be done similarly to the rules.

The output should be the results of queries. This requires a truth value that represents the result of goals, but can also indicate whether results were found in queries. In addition, a list of objects should display the instantiations found for query requests. Based on this summary of the necessary input and output, a UML class diagram can be defined that depicts the necessary functions for creating logic programs in the context of object orientation (see Fig. 1).

| **Inputs** |
|---|
| Creating facts |
| • Predicate with n attributes<br>• Any objects as attributes |
| Creating rules |
| • Rule header as predicate with attributes<br>• Any objects as attributes<br>• Deep condition for rules linked by logical AND/OR<br>• Define variables as attributes |
| Creating queries |
| • Query in the form of a predicate<br>• Any objects as attributes<br>• Define variables as attributes |
| **Outputs** |
| Displaying the results of queries |
| • Display whether result found<br>• Display elaborated mappings |

*Tab. 1: Summary of the inputs and outputs of logic programs*

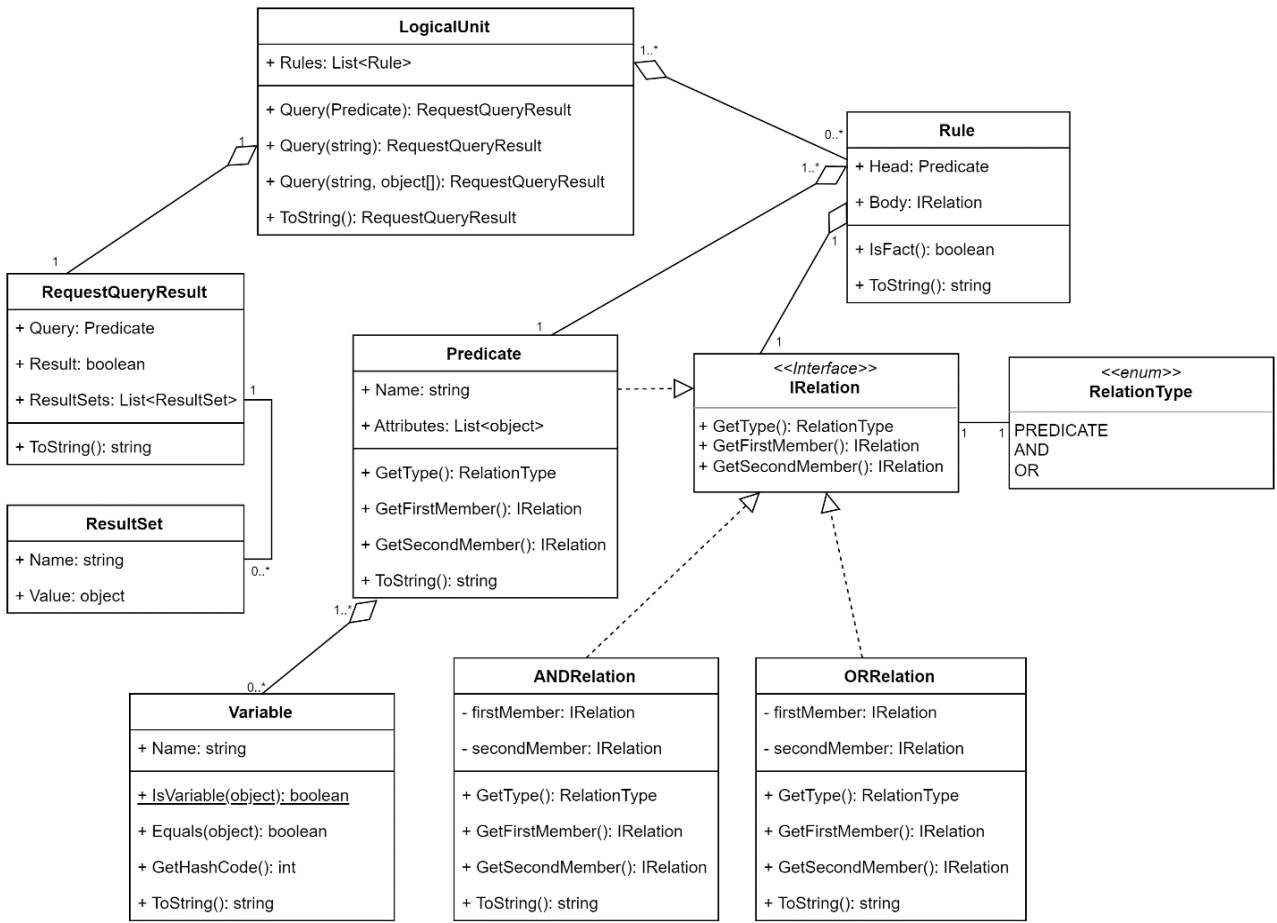

*Fig. 1: Class diagram*

The entry point is an instance of the *LogicalUnit* class, which represents a logic program and in turn can contain instances of the *Rule* class and accepts queries. As described earlier, the program does not distinguish between facts and rules. A rule contains a header, which represents the predicate, and a body, which represents the conditions. If the body is not instantiated, the rule represents a fact, which can be checked using the *IsFact* method.

The head of a rule consists of a predicate, which can be implemented with the Predicate class. A predicate has a name in the form of a string and any number of attributes in the form of objects. If a variable attribute is necessary, the *Variable* class can be used, which only requires a variable name. The class provides the class method *IsVariable(object),* which can be used to check whether any object is a variable.

For the body of a rule it is necessary to enable linking, which is achieved via the interface *IRelation*. The types of relations are defined by the Relation-Type enumeration. The interface has three implementations:

- *ANDRelation*: Has two further *IRelation* objects, which are linked with a logical "AND". The methods *GetFirstMember* and *GetSecondMember* can be used to read the corresponding objects, the method *GetType* returns the type of the link.
- *ORRelation*: Comparable to the *ANDRelation*, but the members are linked with a logical "OR".
- Predicate: A simple predicate also implements the interface and represents the lowest level in a linkage hierarchy. The Predicate is returned both when the *GetFirstMember* method is called and when *GetSecondMember* is called.

Despite the restriction to two members in the interface *IRelation*, more than two objects can be linked by nesting several identical relations within each other. This is possible due to the associativity of the conjunction and disjunction, according to which it holds that $(A \vee B \vee C) = ((A \vee B) \vee C)$ as well as $(A \wedge B \wedge C) = ((A \wedge B) \wedge C)$ [9]. Furthermore, this definition of the interface *IRelation* leads to the fact that a created hierarchy of *IRelations* always represents a binary tree and thus can be traversed like such a tree, since there are unique left and right child nodes.

With this class architecture, a logic program based on objects can now be declared, although these classes are so far only empty shells. As already mentioned, processes and computations are still required, which answer the queries to the *LogicalUnit*. As with the class architecture, the first step is to collect the existing requirements. The following tasks in particular should be mentioned:

- Backward chaining through the knowledge base with backtracking in case of unsuccessful search [2, 10],
- Resolving the hierarchy of possible multiple branched rules,
- Unification variables with constant values [2, 11],
- Resolve clauses [2, 11],
- Processing and return the results.

Based on the query, the knowledge base must be searched and matching rules and facts must be identified and unified or resolved. In the case of rules, the components and any hierarchies of links must be resolved. If the result was compiled, it must be prepared and output. The collection of the requirements makes it possible to model the processes. The designed processes are documented with UML activity diagrams. The diagram in Fig. 2 shows an overview of the entire process, while the detailed process documentation can be reviewed in the appendices 1–3.

When designing the diagrams, the process was divided into three algorithms for reasons of clarity. First, there is a central, recursively executed method that builds and traverses the rule tree based on the query. During the traversal of the knowledge base, it is necessary to search for unifiable rules and output the corresponding mappings, which is the second procedure. Last, a procedure is necessary that reads the value(s) of a variable from the available mappings. This is necessary, because according to the used unification rules cf. the rules of prolog also variables can be mapped to each other, which have the same value afterwards [2]. In order to be able to evaluate all values of a variable, the cross references between variables must also be checked. These three algorithms work together and call each other to produce the final result.

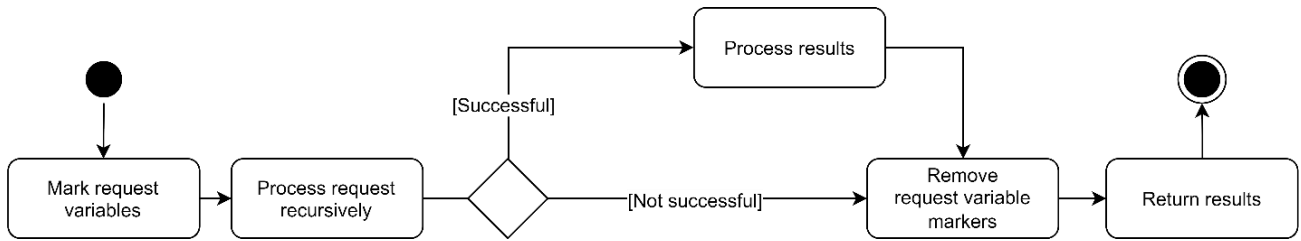

*Fig. 2: Overview resolving process*

At the beginning, it is necessary to label the variables of the query to avoid equating variables that have the same name in the query and in the knowledge base, but are not identical. Then, the query predicate can be processed recursively. This procedure is shown in detail in appendix 1. The recursive approach was chosen because recursion is also being used in the logic programs themselves. By using the same technique, the complexity of the created algorithms is reduced, since the sequences of the logic programs does not have to be compiled in another iteration technique such as loops.

At the beginning of each method call it is checked whether the currently considered *IRelation* contains a predicate, an AND- or an OR-link. In the case of an AND-link, the first part of the link is checked first by calling the base method recursively. If this is successful, the recursive method is called with the second part of the link and the results from the first part until no further result is found. Subsequently, this procedure is repeated until no new result of the first link component can be found. Thus, per iteration, an attempt is made to work out all the corresponding results of the second part with the help of a new result of the first part. If at least one valid combination of the two components has been found, the search is considered successful. In addition, the corresponding unifications for each valid combination are stored and returned at the end of the search.

If an OR-link is considered, the first component of the operation is checked first by calling the method recursively. If this is not successful, the second component is tested. If a valid result could be found for at least one of the two components, the iteration is evaluated as valid and the process is repeated to identify further results. If neither the first nor the second component could provide another valid result, the process is aborted. Analogous to the AND-link, if at least one repetition is valid, the entire search is considered successful, and the unification result is stored and finally returned for each valid iteration round.

For a predicate, the unifiable rules and corresponding mappings are searched. This process is illustrated in detail in appendix 2. In this process, the specifications for the resolution of Horn clauses and the unification of variables are applied. For further information about this technologies please refer to Scott or Robinson [2, 11]. Basically, a rule is accepted as matching if the predicate of the rule head has the same identifier, the number of attributes is identical, and unification is possible. The matching rules are then iterated and checked to see whether they are a rule or a fact, i.e., a rule without a body.

To process a rule, the initial method is recursively called with the rule conditions and the corresponding unifications. If this is successful, the results are saved and the method is called again to identify further results. Otherwise, the loop is aborted and the result is returned.

If the current rule can be identified as fact instead, the target of the current recursion chain has been found. First, it is checked whether the unifications found have been used before, i.e., the unification result is already known. If this is not the case, the search is considered successful and the unification is saved. Furthermore, the unification is marked as used, so that it will not be used again in future runs and only new results will be generated. Since no further call of the initial method takes place at this point, the recursion terminates and the call chain is dismantled.

When the recursion is fully resolved in the sequence, the final result is checked. If this is positive, the results are processed to be returned to the user. Finally, the marking of the query variables is removed and the result is output. The solution finding process is now complete.

In order to be able to integrate the described processes into the presented class architecture, it must be extended again at this point. The extended class diagram can be seen in appendix 4. Four additional classes were designed to implement the solution algorithms, but no changes were made to the existing model classes. The *QueryService* class is initialized with the query predicate and knowledge base when a request is made and is called with the *QueryRequest* method as the entry point. Subsequently, the method *Resolve* is called recursively to traverse the knowledge base as explained in the described flow. For this purpose, an *IRelation*, as well as in each case a list of the new classes *Mapping* and *ProcessedMapping* is passed to the method. The class *Mapping* maps the assignment of an object to a variable, while *ProcessedMapping* is used to note already used unifications per predicates.

In addition, further methods in the class *QueryService* are required for the solution determination. They are called in the method *Resolve*:

- *GetMatchingRules*: The method returns all unifiable rules based on a predicate and the current mappings.
- *Unificate*: This method checks whether two predicates are unifiable assuming the current mappings and returns the result and the corresponding instantiations, if any.
- *GetValues*: This method returns the current values for a variable.
- *Contains*: This method checks whether the current mappings are already contained in the list of *ProcessedMapping*.

Another addition to the class model is the *ExternalMappingService* class, which is responsible for marking the request variables and removing this marking at the end of processing. It also provides the method *GetResultSet*, which is used to process the final result in case of a successful run.

## 3 Programming examples

The introduced architecture was implemented as a C#.NET 6 framework called Logic#. For those interested, the source code can be reviewed and downloaded at GitHub [7]. This chapter uses two examples to show how the architecture can be used to implement a logical program in an object-oriented language, in this case C#. The examples are based on occasional use cases where logic programming is used in practice. This type of programs is often referred to as expert systems, i.e. programs that contain the knowledge of one or more experts and can thus solve queries by logically combining the knowledge [12]. The reason why logical programming is often chosen for this type of program is the proven suitability of the programming language for such problems as already mentioned in the introduction [4].

First, a rudimentary code snippet is shown which accepts three keywords as input and return a string with a help text. In a further stage, a program like this could serve as a chatbot, which, on the basis of keywords in a query, outputs suitable answers. In practice, such bots are used, for example, for customer support or making appointments with doctors, which is why there are large software projects on this topic, such as IBM's Natural Language Processing software "Watson" [13, 14]. All the keywords and sentences in the example code are in a context with an IT service desk, so it could help users to help themselves with computer problems.

```
1  using LogicSharp;
2  using LogicSharp.Models;
3
4  var k1 = new Variable("K1");
5  var k2 = new Variable("K2");
6  var k3 = new Variable("K3");
7  var l = new Variable("L");
8
9  var lunit = new LogicalUnit
10 (
11     new Rule(new Predicate("solution", "computer", "bluescreen", "crashed", "Please restart the PC and report if the problem occurs again!")),
12     new Rule(new Predicate("solution", "computer", "crashed", "again", "Please contact the hotline!")),
13     new Rule(new Predicate("solution", "drucker", "ink", "empty", "Please restart the PC and report if the problem occurs again!")),
14     new Rule(new Predicate("solution", "computer", "hot", "crashed", "Please contact the hotline!")),
15     new Rule(new Predicate("solution", "computer", "shuttered", "liquid", "Please contact the hotline!")),
16
17     new Rule(new Predicate("request", k1, k2, k3, l),
18             new ORRelation(new Predicate("solution", k1, k2, k3, l),
19             new ORRelation(new Predicate("solution", k1, k3, k2, l),
20             new ORRelation(new Predicate("solution", k2, k1, k3, l),
21             new ORRelation(new Predicate("solution", k2, k3, k1, l),
22             new ORRelation(new Predicate("solution", k3, k1, k2, l),
23                         new Predicate("solution", k3, k2, k1, l)))))))
24 );
25
26 Console.WriteLine(lunit.Query(new Predicate("request", "computer", "liquid", "shuttered", l)));
27 Console.WriteLine(lunit.Query(new Predicate("request", "crashed", "hot",
```

Fig. 3: Programming example "Chatbot IT-Service-Desk"

In the lines 4–7 of Fig. 3 variables are declared that can be used in the logic context. The *LogicalUnit* and so the logic program is declared next and supplied with several rules. The first five rules in the lines 11–15 provides the basic knowledge about known keywords and quick help for common issues while the next rule from 17–23 enables the program to find the correct solution no matter what sequence the keywords are. As it can be seen in Fig. 3, all components of this logic program are represented by an object, even the whole logic program as well as all the rules and their parts. The last two lines show how the program can be queried by the corresponding method of the *LogicalUnit* class.

The next example shows a situation where the framework is not used exclusively to solve the query, but is included in a surrounding method, which prepares the input values for the logic queries. This workaround is necessary at this point since the presented approach does not provide direct list processing yet. In order to process a list, nevertheless, the *LogicUnit* was encapsulated. This approach, which is necessary at this point, shows well, how a logical approach can be embedded in normal object-oriented code, if this seems to make sense, without having to do completely logical programming.

Technically, the code snippet shows a medical expert system used by medical professionals in practice [15]. For example, such a system was developed in the 1980s to help physicians diagnose acute abdominal pain. Please note that this example implements only lay medical knowledge and makes no claim to accuracy or completeness of the medical knowledge.

```csharp
1  using LogicSharp;
2  using LogicSharp.Models;
3
4  var s1 = new Variable("S1");
5  var s2 = new Variable("S2");
6  var s3 = new Variable("S3");
7  var d  = new Variable("D");
8  var d2 = new Variable("D2");
9
10 var logicalunit = new LogicalUnit
11 {
12     new Rule(new Predicate("symptoms", "cough", "snuff", "cold")),
13     new Rule(new Predicate("symptoms", "cough", "headache", "cold")),
14     new Rule(new Predicate("symptoms", "cough", "throat_pain", "cold")),
15     new Rule(new Predicate("symptoms", "snuff", "headache", "cold")),
16     new Rule(new Predicate("symptoms", "snuff", "throat_pain", "cold")),
17     new Rule(new Predicate("symptoms", "headache", "throat_pain", "cold")),
18     new Rule(new Predicate("symptoms", "cold", "fever", "influenza")),
19     new Rule(new Predicate("symptoms", "abdominal_pain", "sickness", "gastrointestinal_disease")),
20     new Rule(new Predicate("symptoms", "abdominal_pain", "diarrhea", "gastrointestinal_disease")),
21     new Rule(new Predicate("symptoms", "sickness", "diarrhea", "gastrointestinal_disease")),
22
23     new Rule(new Predicate("diagnosis", s1, s2, d), new ORRelation(new Predicate("symptoms", s1, s2, d),new Predicate("symptoms", s2, s1, d)))
24 );
25
26 List<string> Diagnosis(List<string> liste_symptome)
27 {
28     var diagnosisn = new List<string>();
29     var symptomsToCheck = new List<string>(liste_symptome);
30     var newSymptomsToCheck = new List<string>();
31     do
32     {
33         foreach (var symptom1 in liste_symptome)
34         {
35             foreach (var symptom2 in symptomsToCheck)
36             {
37                 var result = logicalunit.Query(new Predicate("diagnosis", symptom1, symptom2, d)).ResultSets.Select(rs =>
38                                                                                                            rs.Value.ToString()).ToList();
39                 diagnosisn.AddRange(result);
40                 newSymptomsToCheck.AddRange(result);
41             }
42
43         }
44         symptomsToCheck.Clear();
45         symptomsToCheck.AddRange(newSymptomsToCheck.Distinct());
46         newSymptomsToCheck.Clear();
47     }
48     while (symptomsToCheck.Any());
49     return diagnosisn;
50 }
51
52 Console.WriteLine(string.Join(',', Diagnosis(new List<string> { "fever", "snuff", "headache" })));
53 Console.WriteLine(string.Join(',', Diagnosis(new List<string> { "abdominal_pain", "sickness" })));
```

*Fig. 4: Programming example "Medical Expert System"*

The entry point of the program is the method *Diagnosis* in line 26 of Fig. 4, which accepts a list of strings with symptoms. In the further procedure the program prepares the list, so that there are tuples of two symptoms which can be analyzed by the *LogicalUnit* respectively the prepared rules. Since there are recursive diagnoses in the knowledge base, where one diagnosis may be a symptom of another disease, the results of the queries are also added to the list of pending symptom checks and sent to the *LogicUnit* in another iteration. As can be seen in the snippet, it takes surrounding logic to solve this problem, but all the knowledge can be stored centrally in the *LogicUnit*, where a rule can determine if two symptoms are known for a diagnosis.

# 4 Verification

Now that it is clear what an architecture can look like that supports logic in an object-oriented context and it has also been shown how this architecture can work in practice, the question is still open as to whether this approach is useful at all and helps to program better and simpler software. In the course of this question, the presented approach is to be compared in particular with C# without logic components and with the logical language Prolog. In order to answer the question, however, it must be still defined, how programming languages and programming approaches can be compared objectively. For the methodology of such a comparison no references could be found in the literature. The evaluation must therefore be based on the assumption that the suitability of a programming language and the quality of a software artifact in this language correlate positively. This thesis is supported e.g. by an experimental attempt of Moström and Carr, in which it could be shown that the choice of a programming language suitable for the task leads to better programming results [3]. Furthermore, the quality of programs can be objectively measured by using software metrics [16]. For this reason, software metrics should be applied to artifacts in the respective languages and the results can be compared to validate the suitability of the approaches.

For this purpose, test cases are needed, which are implemented, measured, and compared afterwards in all approaches which can be considered. In order to get a first impression, the two programming examples already introduced in chapter 3 will be used, which represent typical application examples of the logic programming. It should be noted that due to the limited scope of the comparison, the results can only be interpreted as a tendency and should not be generalized. The implementations of the application examples in C# and Prolog can be seen in the appendices 5 and 6.

For the selection of software metrics, the "Goal/Question/Metric" paradigm (GQM) shall be applied, which is also referred to in the IEEE standard [16]. When developing the goals, the conventional classification of software metrics into quantitative, qualitative, and complexity metrics should be followed and each of these points should be considered [17].

The following goals and questions were developed for the GQM method:

- Goal: Quantitative evaluation of the different program codes.
    o Question: Which programming variant can be used to solve the expert problems with the least amount of code?
- Goal: Evaluation of the complexity of the different program codes.
    o Question: Which programming variant can be used to program the expert problems most straightforwardly?
- Goal: Qualitative evaluation of the different program codes.
    o Question: How is the overall quality of the individual approaches to be rated?

To answer the first, but also the second question, Halstead's complexity metrics will be used. The Halstead metrics compute on the operands and operators of a program code different complexity-related characteristics, like e.g., the difficulty of understanding the code. However, the metrics also consider attributes such as the length of the program code and thus also represent quantitative properties of the program. The Halstead metrics as representatives of complexity and quantity were chosen because of their widespread use, but it should be mentioned that these metrics, like all metrics, have weaknesses. Thus, particularly complicated program structures are often not adequately represented in the results and only individual code areas can be meaningfully evaluated. [18]

For this reason, to answer the complexity question, the Mc Cabe metric will additionally be considered, since it can identify and evaluate complex structures in contrast to the Halstead metrics. The metric is based on the control flow graph of the measured software and applies to it a function whose image is called a "cyclomatic number" which increases with higher structural complexity. [19]

For the quality determination of the developed software a custom metric is to be defined. The quality of a software can be represented e.g. after Boehm by seven characteristics, whereby these possess again twelve sub-characteristics, which must be evaluated in each case with a metric [20]. Since this amount of evaluation was not possible in the course of this paper, it was decided to evaluate and compare the values obtained from the Halstead and Mc Cabe metrics and to map the overall result. The following values are evaluated per test program:

- lower program scope,
- lower degree of difficulty,
- lower cyclomatic number.

The Halstead metrics for program size and the difficulty are used to integrate one representative each of the complexity and quantity values. The other measured and calculated values of the Halstead metrics are the basis for these values or can be derived directly from them, so the inclusion of other values was not considered to provide any added value. However, the cyclomatic number of the Mc Cabe metric is used to include the structural complexity.

Based on these values, comparisons and evaluations are necessary to map overall quality. In classical quality metrics, individual software artifacts are evaluated, which is why there are no methodologies for comparison here [18]. For example, Boehm's method simply defines target values and calculates deviations, whereas Gilb's method only generates eight values that subsequently need to be interpreted [18, 20]. Since the results of each metric are represented in different number spaces, a classification of the results into an unified number range has to be done, which is called a transformation function [21]. As a basis for this, a function is introduced whose image $Q_M$ is based on the mean value of the result values of a metric and indicates the respective ratio factor, whereby a lower factor represents a better result. A similar approach was also used, for example, by Coleman et al. to standardize various metrics and to evaluate the maintainability of a program code [22].

$$Q_{M(i)} = \begin{cases} \frac{metric_i}{avg(metric)} & \text{if low value better} \\ \\ \frac{1}{\left(\frac{metric_i}{avg(metric)}\right)} & \text{if high value better} \end{cases} \quad (1)$$

Based on the result $Q_M$ of individual metrics, the values $Q_A$ and $Q_P$ can be calculated, which represents the quality of a software artifact or an approach. It applies again that a smaller factor represents a better result.

$$Q_{A(i)} = \frac{\sum Q_{M_i}}{count(Q_M)} \quad (2)$$

$$Q_{P(i)} = \frac{\sum Q_{A_i}}{count(Q_A)} \quad (3)$$

One of the weaknesses of this metric is that it is less meaningful than a full analysis due to the limited number of metrics considered. Since the metric is based on the mean value of the results, the metric does not generate a general quality statement, but can only be considered in the context of the evaluated programs. To calculate the overall quality, the individual values per metric are added together and then divided by the number of metrics. This leads to the metrics being considered equal. If they are to be weighted differently, this is not possible with the calculation method presented. Despite these aspects, the metrics can be used to make a reasonable statement about which approaches tend to achieve better quality values in the development of expert systems, given the specific task.

Based on the definition of software metrics and the implementation, which can be seen in chapter 3 and in appendices 5 and 6, the measurements and calculations can be made. In the following, the results of the software metrics are presented in tabular form. Rounded numbers are represented in the tables in favor of the clarity; however, exact intermediate results were calculated at all times.

| **Halstead metrics** | | IT-Service-Desk | | | Medical Expert System | | |
|---|---|---|---|---|---|---|---|
| | | C# | Prolog | Logic# | C# | Prolog | Logic# |
| $n_1$ | Count operators | 21 | 6 | 8 | 29 | 9 | 14 |
| $n_2$ | Count operands | 47 | 9 | 30 | 54 | 18 | 54 |
| $N_1$ | Total operators | 218 | 84 | 186 | 289 | 140 | 302 |
| $N_2$ | Total operands | 169 | 70 | 117 | 145 | 102 | 166 |
| N | Program length | 387 | 154 | 303 | 434 | 242 | 468 |
| n | Vocabulary size | 68 | 15 | 38 | 83 | 27 | 68 |
| U | Program scope | 2,356 | 602 | 1,590 | 2,767 | 1,151 | 2,849 |
| S | Difficulty | 38 | 23 | 16 | 39 | 26 | 22 |
| A | Workload | 88,946 | 14,039 | 24,806 | 10,7725 | 2,9342 | 61,305 |
| D | Duration in seconds | 4,941 | 780 | 1378 | 5,985 | 1,630 | 3,406 |

*Tab. 2: Halsteadt metrics*

For the Halstead metrics in Tab. 2, the developed programs were measured and the number of different operators and operands and the respective total numbers of these were noted. Subsequently, the individual values of the metric were calculated.

| **Mc Cabe metric** | | IT-Service-Desk | | | Medical Expert System | | |
|---|---|---|---|---|---|---|---|
| | | C# | Prolog | Logic# | C# | Prolog | Logic# |
| e | Total edges | 6 | 1 | 1 | 7 | 1 | 7 |
| n | Total nodes | 5 | 2 | 2 | 5 | 2 | 6 |
| v(G) | Cyclomatic number | 3 | 1 | 1 | 4 | 1 | 3 |

*Tab. 3: Mc Cabe metric*

To count the edges and nodes for the Mc Cabe metric in Tab. 3, control flow graphs were created and the measurements were entered into the table. Then the cyclomatic number was calculated according to the formula from the Mc Cabe metric. The designed control flow graphs are shown in appendix 7.

| **Quality metric** | | IT-Service-Desk | | | Medical Expert System | | |
|---|---|---|---|---|---|---|---|
| | | C# | Prolog | Logic# | C# | Prolog | Logic# |
| U | Program scope | 2,356 | 602 | 1,590 | 2,767 | 1,151 | 2,849 |
| S | Difficulty | 38 | 23 | 16 | 39 | 26 | 22 |
| v(G) | Cyclomatic number | 3 | 1 | 1 | 4 | 1 | 3 |
| $Q_{M(U)}$ | Factor program scope | 1.55 | 0.40 | 1.05 | 1.23 | 0,51 | 1.26 |
| $Q_{M(S)}$ | Factor difficulty | 1.48 | 0.91 | 0.61 | 1.36 | 0,89 | 0.75 |
| $Q_{M(v(G))}$ | Factor cyclomatic number | 1.80 | 0.60 | 0.60 | 1.50 | 0.38 | 1.13 |
| $Q_{A(i)}$ | Factor artefact | 1.61 | 0.64 | 0.75 | 1.36 | 0.59 | 1.05 |

*Tab. 4: Quality metric*

For the quality metric in Tab 4, the results from Tab. 2 and Tab. 3 were used and the factors were calculated according to the specified formulas. Based on the values, the total factors $Q_P$ per approach can now be calculated:

- $Q_{P(C\#)} = 1.49$,
- $Q_{P(Prolog)} = 0.61$,
- $Q_{P(Logic\#)} = 0.90$.

The results of the metrics confirm the assumption from the introduction that logical programming, in this case with the representative Prolog, offers a good basis for the definition of expert systems. Thus, Prolog could achieve the best results in nearly all values and with both programs. The only exception is the difficulty value of the Halstead metrics, which can be explained by the fact that this evaluates the difficulty independently of the extent of the artifact. Although the Prolog variant has a lower overall complexity, which is shown by the other complexity values, due to its significantly lower scope this low overall complexity is distributed over less code and is therefore rated as harder to understand. As mentioned, the complexity in all other values is nevertheless lower than in the other approaches. It can be said that Prolog also scores best in complexity. This also results in an overall quality score of 0.61 in the quality metric, which is the best result in the field. The second best quality score of 0.90 was achieved by the Logic# implementation, which was inferior to the Prolog implementation in each case, but still produced above-average results due to the score < 1. In particular, the scope of the Logic# versions were higher in each case in comparison to the corresponding Prolog counterpart. This can be attributed primarily to the fact that C#-typical language constructs such as the "new" keyword were necessary for declaring new objects, whereas this is not required for Prolog. Furthermore, in the case of the medical expert system, it was necessary to first prepare the passed list in upstream code, since no direct list processing was provided for the Logic# prototype, while Prolog already supports this natively. For the cyclomatic number, Logic# was able to achieve the best value of 1 in the chatbot test case, on par with Prolog. This is due to the fact that no sequential processing had to be defined in the code in either program, since this is taken over by the interpreter and the framework, respectively. However, since preparatory work was necessary in the Logic# solution for the medical expert system, a significantly worse result was achieved here than with Prolog. In comparison to the C# implementation, which achieved an overall score of 1.49, well above 1 and thus the worst result, it can be seen that a reduced complexity could be achieved in particular through the developed Logic# framework. Thus, although a marginally lower scope was achieved for the chatbot system, the workload $A$ and difficulty $S$ could be reduced many times over by using the framework. These significant results are also supported by the cyclomatic number, which is three times higher for C#. In the case of the medical expert system, these significant improvements could not be achieved, because, as already described, preparatory work was necessary. As a result, a slightly higher amount of code was required here for the Logic# solution, but lower complexity was still shown in both the Halstead and Mc Cabe metrics.

In summary, the Prolog development performed best in all three objectives and questions of the GQM method. However, it was shown that with the help of the addition of the developed framework to C#, significant relief in terms of program complexity is possible. For this reason, the Logic# solution scored second best in this aspect, while native C# could only achieve the last place. In the area of code quantity, C#, and Logic# are about equal and share the second place. The developed quality metric represents the overall quality and showed better values for Logic#, which is why this approach receives the second place, while C# also brings up the rear here.

# 5 Conclusion

This paper addressed the porting of logical programming to the object-oriented language C#. Current implementations of logic in C# or other object-oriented languages, where they are mostly represented by Prolog, include full-fledged interpreters based on the respective object-oriented language and in which Prolog can be programmed or require interfaces to other programs. In contrast, this paper shows how logical programming can be integrated directly into C# code. The presented software architecture contains a class concept and the necessary algorithmic processes that are required for this purpose.

Based on the architecture, an exemplary C# framework was realized, and implementation examples were presented. Thus, it could be shown, how typical functionalities of logic languages can be implemented in an object-oriented language, without contradicting the thought of the object orientation. For the use only a program library must be imported, as this is usual for all current Frameworks e.g., for LINQ and the functional programming. Subsequently, the classes and functionalities of Logic# can be used by programmers for their own programs or program parts. The work showed how chatbots or medical expert systems can be implemented with this approach.

During the development of the example programs, however, also weaknesses of the approaches became apparent. For example, no list processing was implemented because this is already integrated in object-oriented languages. However, it became clear that there are significant differences between the logic and the C# list processing, why partially awkward workarounds would have to be programmed, as this was evident in programming example two. A further conceivable extension of the Frameworks for the significant extension of the function range, would be the opening of the interface *IRelation* around a further type, in which an arbitrary method can be encapsulated. If this method is passed the current mappings and returns a new list of mappings, user-defined processes such as arithmetic calculations can be implemented during the unification process. Such a functionality would be e.g., necessary for the implementation of a faculty function. Thus, the current functionality of the frameworks does not offer the possibilities to perform a multiplication within the logic program and to assign the result to a variable. Despite these extension needs and possibilities, the current prototype already shows a preview of how logic programming is possible in an object-oriented context.

For the validation of the approach, a comparison between C#, Logic# and Prolog was defined using software metrics, a balanced selection of metrics was made and two example programs from the field of expert systems were defined. Subsequently, the test programs were implemented, measured, and evaluated in the different paradigms.

This experimental setup came to the result that advantages could be reached by a logic implementation in C# if expert systems are to be implemented. Especially in the area of complexity, the Logic# implementations showed significantly better results than the C# variants. Even though the amount of code required remained about the same, lower Logic# complexity resulted in better overall program quality, implying better pragmatics. Despite this improvement over native C#, it became apparent that further progress was possible, as Prolog was able to achieve even better results in all values, sometimes outperforming the results of the two competitors by a multiple. This could be attributed thereby above all to not changeable language constructs of C# and the object orientation. In addition, the already described improvements such as an integrated list processing can help to achieve further facilitations for the expert system development in Logic# and other frameworks based on the introduced approach.

At this point it is again to be pointed out that these results should not be generalized for all expert systems, because, in this paper only two test programs were analyzed. In addition to the objective metrics considered, there are other objective measurements that could not be captured by the static metrics, such as the runtime of a program, which could determine whether a program variant is appropriate for the current application.

However, subjective influencing factors are also conceivable, which cannot be quantified exactly and were therefore excluded in this work. Thus, experienced C# developers can solve a problem quite faster in C# than inexperienced colleagues in Prolog, although according to Halstead metric the C# variant requires a much higher workload. In the industrial environment, there are sometimes specifications as to which languages and frameworks may be used in order to avoid uncontrolled growth. This shows that the choice of the right paradigm and programming language depends on a variety of factors and thus on the individual case. Nevertheless, the results of this work indicate that for problems from the field of expert systems or those that are fundamentally similar to complex queries on a knowledge base, the use of logic programming is also useful in the object-oriented environment and therefore should always be considered as an option.

# Appendices

## Appendix 1: Activity diagram – Recursive main flow

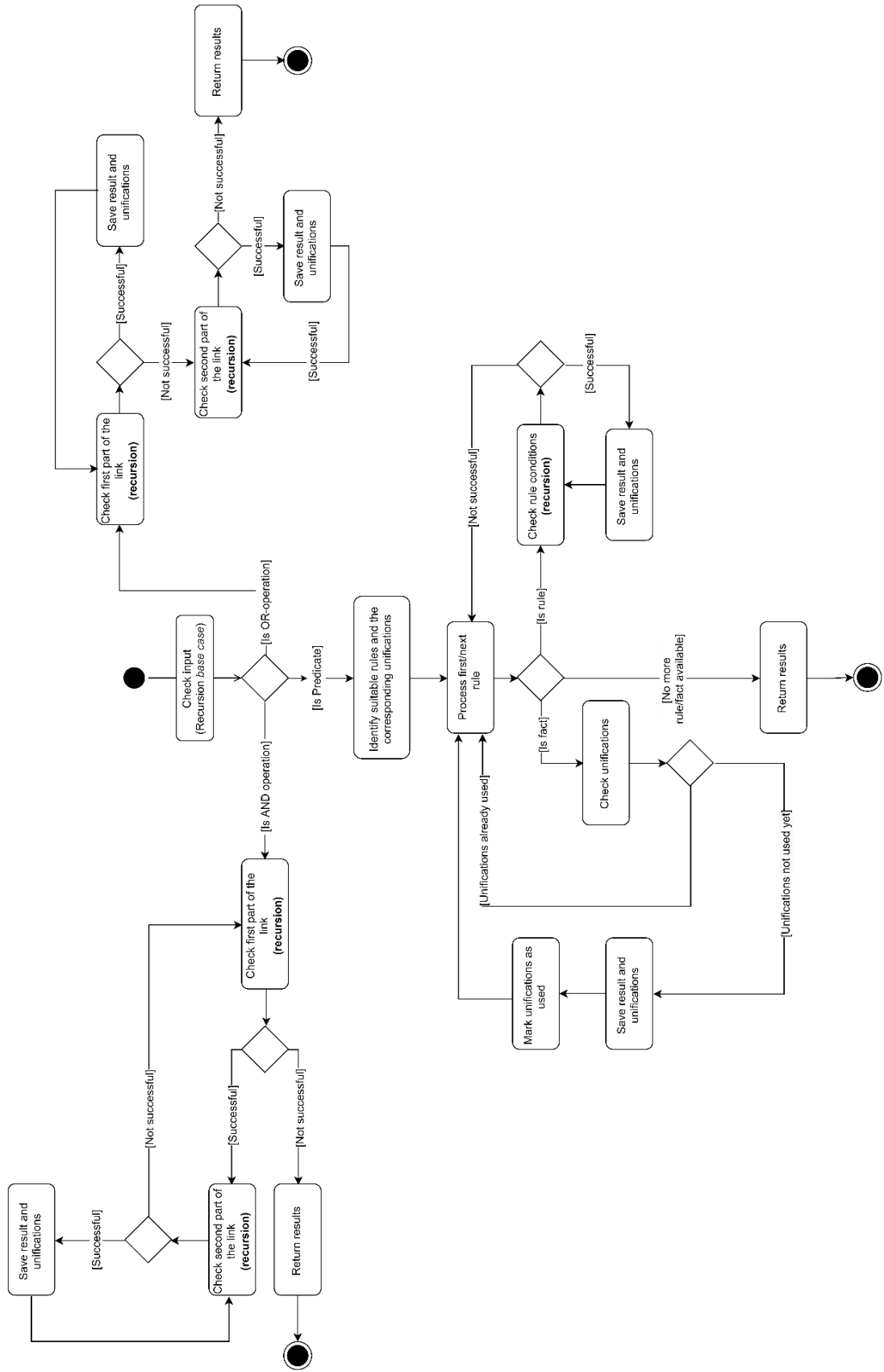

## Appendix 2: Activity diagram – Rule Search & Unification

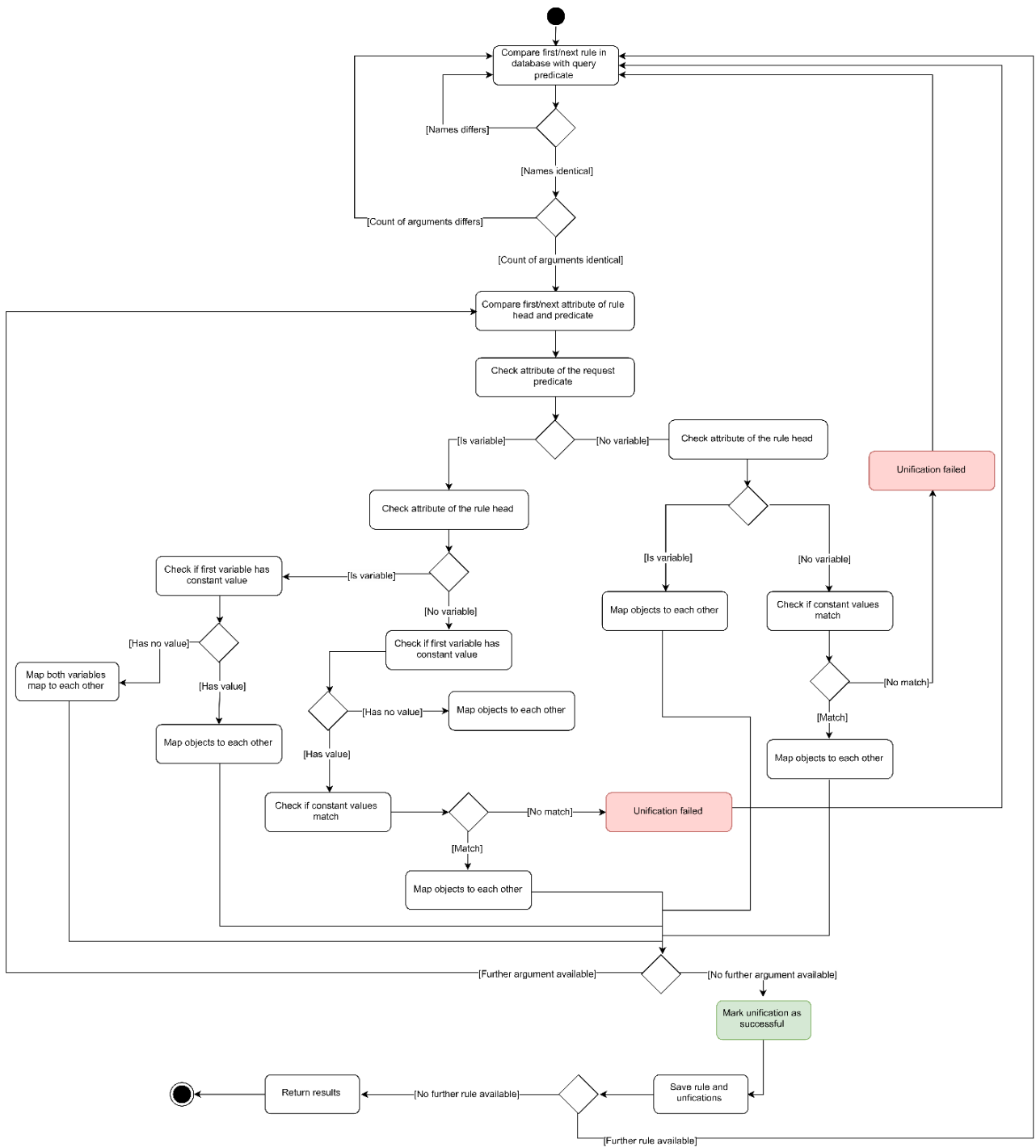

**Appendix 3: Activity diagram – Determination of variable value**

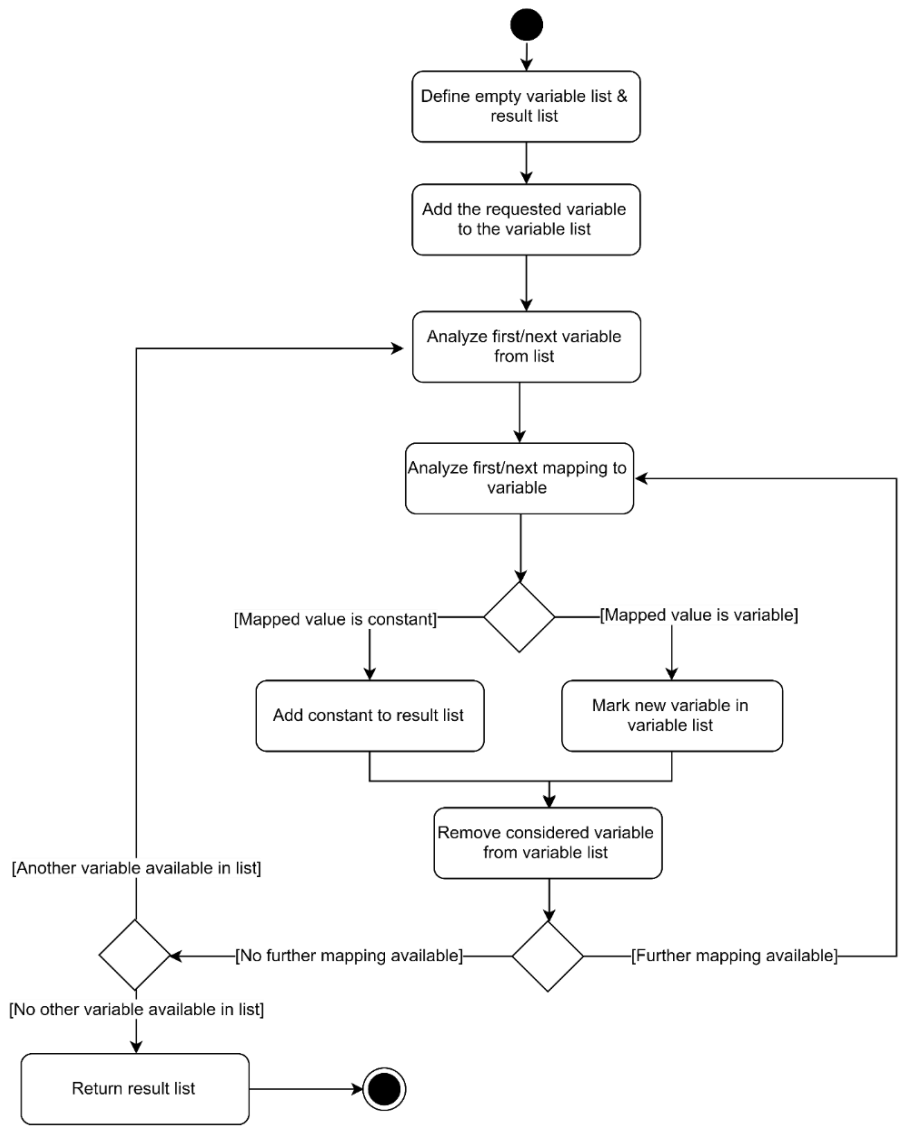

## Appendix 4: Extended class diagram

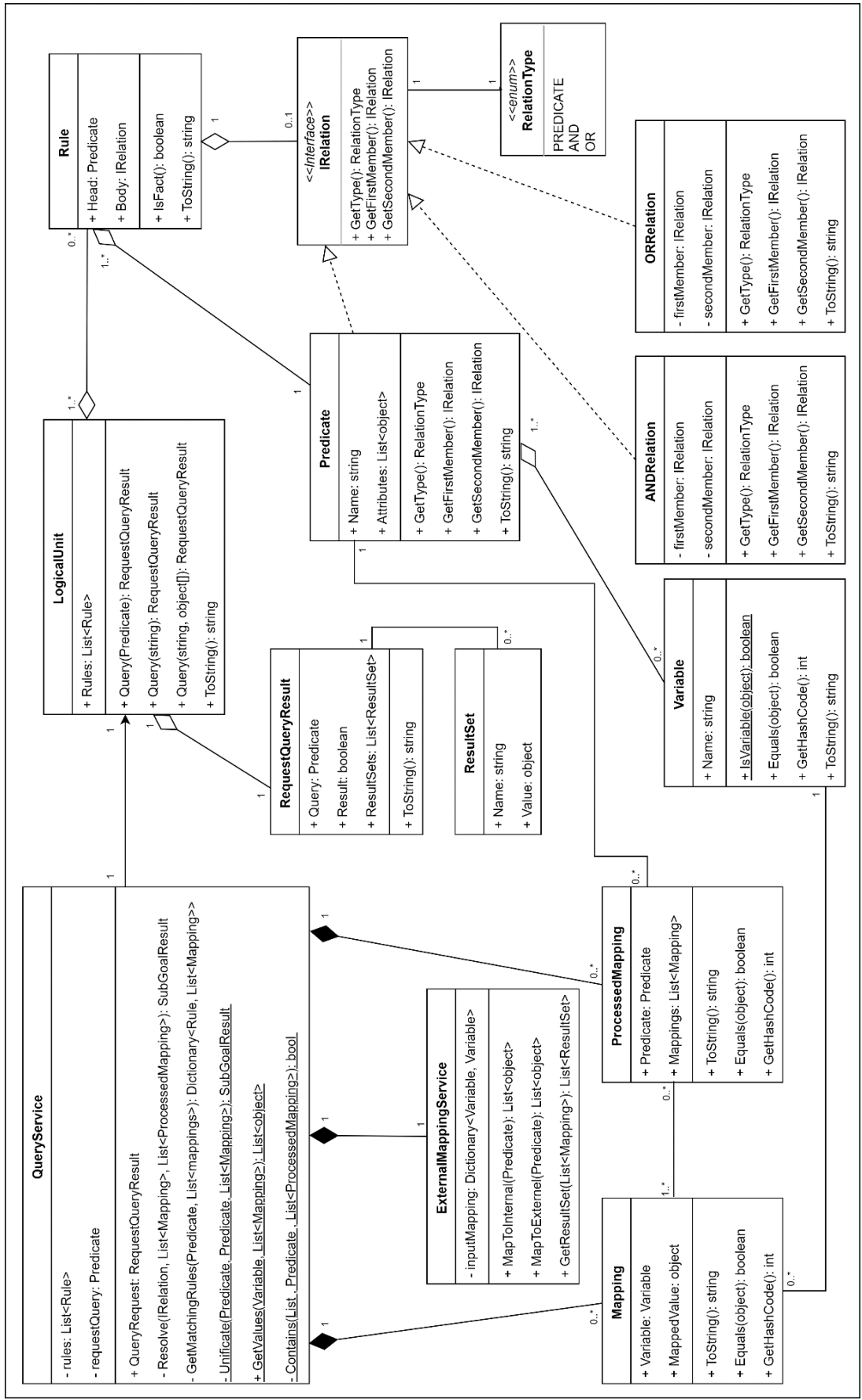

## Appendix 5: Test program "Chatbot IT Service Desk"

### SWI-Prolog implementation

```prolog
solution(computer,bluescreen,crashed,"Please restart the PC and report if the problem occurs again!").
solution(computer,crashed,erneut,"Please contact the hotline!").
solution(printer,ink,empty,"Please ask your secratariat for a spare cartridge!").
solution(computer,hot,crashed,"Please contact the hotline!").
solution(computer,shuttered,liquid,"Please contact the hotline!").

request(K1,K2,K3,L):-solution(K1,K2,K3,L);
                    solution(K1,K3,K2,L);
                    solution(K2,K1,K3,L);
                    solution(K2,K3,K1,L);
                    solution(K3,K1,K2,L);
                    solution(K3,K2,K1,L).

request(computer,liquid,shuttered,L).
request(crashed,hot,computer,L).
```

### C# implementation

```csharp
namespace main_namespace
{
    class Program
    {
        static void Main(string[] args)
        {
            Console.WriteLine(string.Join('\n', Request("Computer", "Liquid", "Shuttered").Select(a => a.ToString())));

            Console.WriteLine(string.Join('\n', Request("Crashed", "Hot", "Computer").Select(a => a.ToString())));
        }

        static List<Solution> solutions = new List<Solution>() {
            new Solution(){ Keywords = new List<string>(){ "Computer", "Bluescreen", "Crashed" } , Helptext ="Please restart the PC and report if the problem occurs again!" },
            new Solution(){ Keywords = new List<string>(){ "Computer", "Crashed", "Again" } , Helptext ="Please contact the hotline!" },
            new Solution(){ Keywords = new List<string>(){ "Printer", "Ink", "Empty" } , Helptext ="Please ask your secratariat for a spare cartridge!" },
            new Solution(){ Keywords = new List<string>(){ "Computer", "Hot", "Crashed" } , Helptext ="Please contact the hotline!" },
            new Solution(){ Keywords = new List<string>(){ "Computer", "Shuttered", "Liquid" } , Helptext ="Please contact the hotline!" },
        };

        static List<Solution> Request(string keyword1, string keyword2, string keyword3)
        {
            var matchingSolutions = new List<Solution>();

            for (int i = 0; i < solutions.Count; i++)
            {
                if (solutions[i].Keywords.Contains(keyword1) &&
                    solutions[i].Keywords.Contains(keyword2) &&
                    solutions[i].Keywords.Contains(keyword3))
                {
                    matchingSolutions.Add(solutions[i]);
                }
            }
            return matchingSolutions;
        }

        public class Solution
        {
            private List<string> keywords;
            private string helptext;

            public List<string> Keywords { get => keywords; set => keywords = value; }
            public string Helptext { get => helptext; set => helptext = value; }

            public override string ToString()
            {
                return Helptext;
            }
        }
    }
}
```

## Appendix 6: Test program "Medical Expert System"

### SWI-Prolog implementation

```prolog
1  symptoms(cough,snuff,cold).
2  symptoms(cough,headache,cold).
3  symptoms(cough,throat_pain,cold).
4  symptoms(schnupfen,headache,cold).
5  symptoms(schnupfen,throat_pain,cold).
6  symptoms(headache,throat_pain,cold).
7  symptoms(cold,fever,influenza).
8  symptoms(abdominal_pain,nausea,gastrointestinal_disease).
9  symptoms(abdominal_pain,diarrhea,gastrointestinal_disease).
10 symptoms(nausea,diarrhea,gastrointestinal_disease).
11 
12 diagnosis(S1,S2,D):-symptoms(S1,S2,D);symptoms(S2,S1,D).
13 diagnosis(S1,[S2|SL],D):-diagnosis(S1,S2,D);diagnosis(S1,SL,D);diagnosis(S2,SL,D).
14 diagnosis([S1|SL],D):-diagnosis(S1,SL,D).
15 diagnosis([S1|SL],D):-diagnosis(S1,SL,DL),( diagnosis (DL,S1,D); diagnosis (DL,SL,D)).
16 
17 diagnosis([fever, snuff,headache],D).
18 diagnosis([abdominal_pain,sickness],D).
```

### C# implementation

```csharp
1  namespace main_namespace
2  {
3      class Program
4      {
5          static void Main(string[] args)
6          {
7              Console.WriteLine(string.Join(',', Diagnosis(new List<string> { "Fever", "Snuff", "Headache" })));
8              Console.WriteLine(string.Join(',', Diagnosis(new List<string> { "Abdominal_pain", "Sickness" })));
9          }
10 
11         static List<Disease> diseases = new()
12         {
13             new Disease { Name = "Cold", Symptome = new() { "Cough", "Snuff", "Headache", "Throat_pain" }, MinSymptoms = 2 },
14             new Disease { Name = "Influenza", DependingDisease = "Cold", Symptome = new() { "Fever" }, MinSymptoms = 1 },
15             new Disease { Name = "Gastrointestinal disease", Symptome = new() { "Abdominal_pain", "Sickness", "Diarrhea" }, MinSymptoms = 2 }
16         };
17 
18         static List<Disease> Diagnosis(List<string> liste_symptome, Disease? pre_disease = null)
19         {
20             var possibleDiseases = diseases.Where(k => (pre_disease == null || k.DependingDisease == pre_disease.Name) &&
21                                       k.Symptome.Intersect(liste_symptome).Count() >= k.MinSymptoms).ToList();
22 
23             var diseasesTodo = new List<Disease>(possibleDiseases);
24             while (diseasesTodo.Count > 0)
25             {
26                 var diseases = Diagnosis(liste_symptome, diseasesTodo[0]);
27                 var newDiseases = possibleDiseases.Intersect(diseases);
28                 diseasesTodo.AddRange(newDiseases);
29                 possibleDiseases.AddRange(newDiseases);
30                 diseasesTodo.RemoveAt(0);
31             }
32             return possibleDiseases;
33         }
34     }
35     public class Disease
36     {
37         private string name;
38         private List<string> symptoms;
39         private int minSymptoms;
40         private string dependingDisease;
41 
42         public string Name { get => name; set => name = value; }
43         public List<string> Symptome { get => symptoms; set => symptoms = value; }
44         public string DependingDisease { get => dependingDisease; set => dependingDisease = value; }
45         public int MinSymptoms { get => minSymptoms; set => minSymptoms = value; }
46 
47         public override bool Equals(object? obj)
48         {
49             return obj is Disease && ((Disease)obj).Name == Name;
50         }
51 
52         public override string ToString()
53         {
54             return Name;
55         }
56     }
57 }
58
```

**Appendix 7: Control flow graphs**

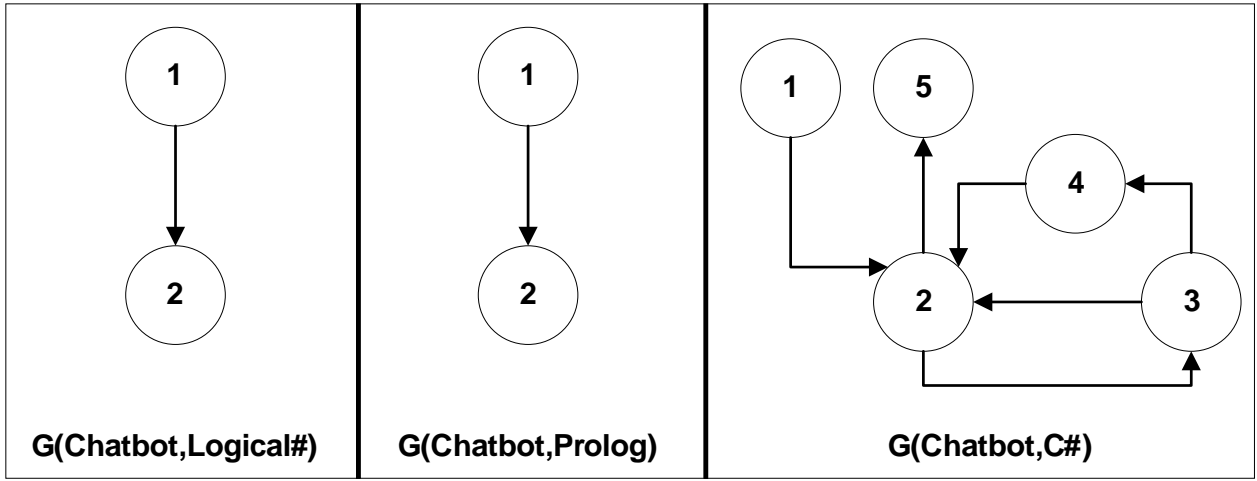

G(Chatbot,Logical#)       G(Chatbot,Prolog)       G(Chatbot,C#)

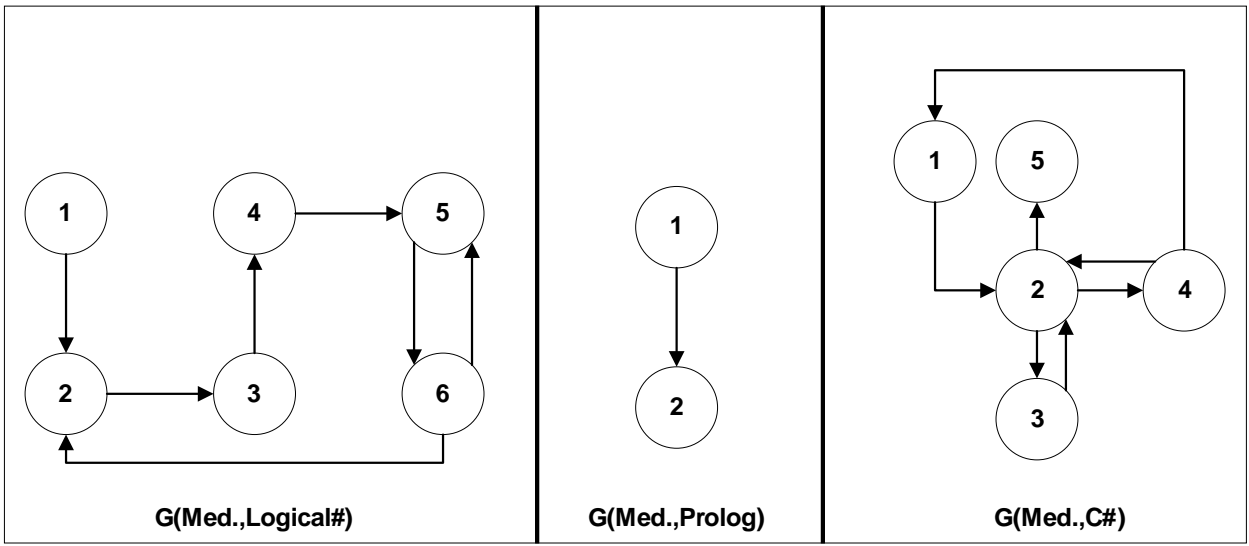

G(Med.,Logical#)       G(Med.,Prolog)       G(Med.,C#)